\documentclass[twoside,leqno,twocolumn]{article}

\usepackage[letterpaper]{geometry}

\usepackage{ltexpprt}
\usepackage{hyperref}

\usepackage{balance}

\usepackage{epsfig}
\usepackage{amsmath}
\usepackage{color}
\usepackage{systeme}
\usepackage{algorithm}
\usepackage{graphicx}
\usepackage{subcaption}
\usepackage{float}
\usepackage[noend]{algpseudocode}
\usepackage{textcomp}
\usepackage{float}
\usepackage{algorithm}
\usepackage{paralist}
\usepackage{booktabs}
\usepackage{amsfonts}

\newcommand{\glmucb}{\texttt{GLM-GT-UCB}}

\newcommand{\linucb}{\texttt{LogNorm-LinUCB}}

\DeclareMathOperator*{\argmax}{argmax}
\DeclareMathOperator*{\argmin}{argmin}
\newtheorem{problem}{Problem}

\newcommand{\blue}[1]{\textcolor{blue}{#1}}

\newcommand\eat[1]{}

\begin{document}

\newcommand\relatedversion{}

\title{Contextual Bandits for Advertising Campaigns: \\ A Diffusion-Model Independent Approach\\
    (Extended Version)}

\author{Alexandra Iacob
\thanks{iacob@lisn.fr, LISN, CNRS, University Paris-Saclay}
\and
Bogdan Cautis
\thanks{cautis@lisn.fr, LISN, CNRS, University Paris-Saclay}
\and 
Silviu Maniu
\thanks{maniu@lisn.fr, LISN, CNRS, University Paris-Saclay}
}

\date{}

\maketitle







\begin{abstract}  Motivated by scenarios of information diffusion and advertising in social media, we study  an \emph{influence maximization} problem in which little is assumed to be known about the diffusion network or about the model that determines how information may propagate. In such a highly uncertain environment, one can focus on \emph{multi-round diffusion campaigns}, with the objective to maximize the number of distinct users that are influenced or activated, starting from a known base of few influential nodes. \eat{(the potential spread seeds), a.k.a. \emph{influencers}}  During a campaign, spread seeds are selected sequentially at consecutive rounds, and  feedback is collected in the form of the activated nodes at each round. A round’s impact (reward) is then quantified as the number of \emph{newly activated nodes}. \eat{, i.e., those that were not already activated at previous rounds.} Overall, one must maximize the campaign’s total spread, as the sum of rounds' rewards.  In this setting, an explore-exploit approach could be used to learn the key underlying diffusion parameters, while running the campaign. We describe and compare two methods of \emph{contextual multi-armed bandits}, with \emph{upper-confidence bounds} on the remaining potential of influencers, one using a generalized linear model and the Good-Turing estimator for remaining potential (\glmucb), and another one that directly adapts the LinUCB algorithm to our setting (\linucb).  We show that they outperform baseline methods using  state-of-the-art ideas, on synthetic and real-world data, while at the same time exhibiting different and complementary behavior, depending on the scenarios in which they are deployed.  \end{abstract}

\section{Introduction}
Social media advertising is a booming domain,  gradually replacing advertising over traditional channels. \eat{(TV, radio, print, mail)} It is enabled by the highly effective word-of-mouth mechanisms that are embedded in social applications, such as likes, shares, reposts, or notifications. \eat{and there are currently more than 3B social media users} Social networking applications are therefore an unprecedented medium for advertising, be it with a commercial intent or not, as products, news, ideas,  political manifests, etc., can propagate easily to a large yet well-targeted audience.  

\eat{But the interest of social media for marketers is not only that it enables to easily and rapidly reach a large user base, but also and foremost that it brings credibility to the messages that are being conveyed. Indeed, there are many studies showing that people are more inclined to pay attention to a message or referral coming from a known individual, e.g., a friend or an influencer whom she follows \cite{Referral1, Referral2}.}

Motivated by advertising in social media, the class of algorithmic problems under the generic name of \emph{influence maximization} (IM) \cite{kempe2003maximizing} encompasses all scenarios that aim to maximize the spread of information in a diffusion network, by identifying the most influential nodes from which the diffusion of specific message should start.  IM  mirrors an increasingly used and highly effective form of marketing in social media, targeting a sub-population of \emph{influential people}, instead of all users of interest, known as \emph{influencer marketing} \cite{InfluencerMarketing}.  

IM usually has as objective the \emph{expected spread} under a stochastic diffusion model, which describes a diffusion as a probabilistic process.  The seminal work of \cite{kempe2003maximizing} introduced two such \eat{stochastic, discrete-time diffusion} models -- Linear Threshold (LT) and Independent Cascade (IC) -- which have been  adopted by most of the literature \eat{that followed} (see the survey of \cite{8295265}). Such models rely on diffusion graphs with edges weighted by a spread probability.

As selecting the seed nodes maximizing the expected spread is NP-hard under such diffusion models, approximation algorithms that exploit the objective's monotonicity and sub-modularity have been studied extensively, yet scaling IM to realistic graphs remains difficult. While most of the IM literature focuses on improving  efficiency and scalability (see benchmarks  \cite{DBLP:conf/edbt/0001GR19,DBLP:conf/sigmod/AroraGR17}), other major obstacles have limited the practical impact of this research. First, it is hard to obtain meaningful influence probabilities, as it is hard and data-intensive to learn them from \eat{a history of}past diffusions. \eat{\emph{(if at all available)}} \cite{DBLP:journals/datamine/HuCCCGH19,DBLP:journals/tkdd/Gomez-RodriguezLK12,DBLP:conf/aistats/DuSWZ13}.  
Also, the effectiveness of most IM algorithms depends on diffusion models and their key parameters --  whether known or learned in online manner -- aspects which are most often hard to align with real-life diffusion dynamics. It is commonly agreed that such parametric diffusion models represent elegant yet coarse interpretations of a reality that is complex and uncertain.

For these reasons, the focus of the IM literature has shifted recently towards \emph{online} and \emph{diffusion-model independent} methods \cite{lagree2018algorithms, DBLP:conf/icml/VaswaniKWGLS17, DBLP:conf/kdd/LeiMMCS15} where, during a \emph{multi-round influence campaign}, a learning agent sequentially selects at each round seeds from which a new diffusion of the campaign's message is initiated and observed in the network. A round's feedback is then used to update the agent’s knowledge. To balance between exploration (of uncertain aspects of the diffusion environment) and exploitation (e.g., focusing on the most promising seeds), such methods rely on \emph{multi-armed bandits}.

Our work follows in this path, as we study an IM problem in which the diffusion topology, the influence probabilities, and the model that determines how information may spread are all assumed to be unknown. Instead, what is known are the potential spread initiators, a set of few influential nodes called hereafter the \emph{influencers}. In such a highly uncertain environment, under \emph{budget limitations} (number of seedings and number of rounds), the campaign aims to maximize the number of \emph{distinct} users that are influenced or activated, starting from the influencers. Seeds are selected sequentially (at each round) among the influencers, and an influencer may be re-seeded, i.e., selected at multiple rounds. After a round's diffusion, the assumed feedback are all the activated nodes from that round, i.e., only the diffusion's effects are observed (the \emph{who}), but not their causes (the \emph{why}). Generically, this feedback is used to refine the estimations for the influencers' remaining spread potential, which will guide future seeding decisions. Matching the overall objective, a round’s \emph{reward} is the number of \emph{newly activated nodes}, i.e., those that were not already activated at previous rounds. The campaign's objective is to maximize the sum of  rounds' rewards.  

Our problem setting is directly inspired by influencer mar\-ke\-ting scenarios where marketers have only access to a few influencers who can spread information,  where the only feed\-back that can be realistically gathered are activations (e.g., who purchased or subscribed), and where the goal is to maximize the number of distinct activated users (instead of the number of activations).\eat{ More generally, the interest of algorithms for this problem is threefold: (i) they can be used in scenarios with high uncertainty regarding the diffusion environment; (ii) by being diffusion-model independent and working with few assumptions of its own, they are generic and adaptive, with a broad range of applicability, and (iii) by focusing on influencers – the only aspect of the diffusion environment that is assumed to be known – they have a clear algorithmic interest, as the only parameters to be learned are tied to influencers.}  

We follow a \emph{contextual multi-armed bandits} \cite{DBLP:journals/ftml/Slivkins19} approach,  assuming that contextual information is known and exploitable in the sequential learning process, as features of influencers or of the information being diffused. The intuition is that within a campaign, whose overall goal is to get a specific ``message’’ to as many users as possible, the ways in which that message may be formulated, presented, or diffused may vary from round to round, and such contextual variations will lead to different propagation dynamics. E.g., the campaign’s message may be a political manifesto, while the to-be-diffused items may pertain to different aspects thereof, to connections with societal issues, may be framed in news, op-eds, data analysis, multimedia content, etc.    

\vspace{1mm}

\noindent \textit{Contribution.}
We propose two UCB-like algorithms, \glmucb\ and \linucb,  for the problem of selecting influencers in advertising campaigns, where newly activated nodes make up the reward. They follow \emph{optimism in the face of uncertainty} in sequential learning \cite{ DBLP:journals/ftml/BubeckC12},  deriving an upper-confidence bound on the estimator of the remaining spread potential of each influencer. This enables to alternate in a principled way between explore and exploit steps when taking seeding decisions at the campaign's rounds.  Our solutions are diffusion-model agnostic and follow different assumptions on the rewards distribution:  Poisson for \glmucb, log-normal  for \linucb.\eat{-- but do not assume any diffusion model.} \glmucb~ uses a Good-Turing estimator \cite{good1953population, DBLP:journals/jmlr/BubeckEG13} for new activations, to which it applies an external factor function modeling an influencer's fatigue (diminishing returns) and potential in a given context. The parameter of the external factor is assumed to be a linear combination of the context and an unknown feature vector learned through linear regression. \linucb~ assumes a linear structure for the scale of rewards, estimated by the inner product of the context and the influencer's learned feature vector. \eat{In both solutions, decisions are taken based on upper-confidence bounds.}We experimented with synthetic and real-world data, comparing  with state-of-the-art solutions adapted to our problem. The experiments show that our methods successfully learn from the available side-information and achieve higher cumulative rewards. These results are complemented by theoretical \emph{regret} guarantees for a \linucb\ variant  that learns from independent samples. 

\section{Main Related Work}
   The work of \cite{li2017provably} proposed a solution for the generalized linear contextual bandit problem, earlier considered also in a practical scenario of news recommendation \cite{DBLP:conf/www/LiCLS10}. The solution is based on the work of \cite{filippi2010parametric} -- considering non-linear rewards for the MAB problem --  and it improves it by adapting the algorithm of \cite{auer2002using} to use MLE for estimating the unknown parameters, and uses the same approach to create the independent samples.
    
   In \cite{NIPS2017_7137debd}, the authors proposed an UCB-based algorithm, \textit{IMLinUCB} for the online influence maximization problem in social networks. They assume that the diffusion of information follows the independent cascade (IC) edge semi-bandit model\eat{, based on the IC model of \cite{kempe2003maximizing}}. The algorithm selects multiple influencers per round without suffering from an exponential increase in the combinatorial action space due to the cardinality of the source node set. Efficiency is obtained through the linear generalization of a probability weight function that yields the activation probabilities. 
  
    The cumulative regret bounds for \textit{IMLinUCB} are topology dependent; this is also confirmed by the experiments performed on different types of graph topologies.

     In \cite{levine2017rotting}, the authors consider the weariness of an influencer's effectiveness over time and introduce the so called \emph{rotting bandits}. It assumes that the expected reward decays as a function of the number of times an arm has been selected, thus the optimal policy being one of choosing different arms. Our problem bears similarities to the non-parametric rotting bandit problem of \cite{levine2017rotting}, as we also do not make assumptions about the structure of the reward, but only about its non-increasing nature in the number of selections. To this end, \cite{levine2017rotting} proposed the Sliding-Window Average (\textit{SWA}) algorithm. In the initialization phase, each arm is chosen for a fixed number of times, and for the rest of the ``campaign'' their empirical average reward is adjusted by a given quantity. SWA is thus able to detect early the significantly sub-optimal arms, while preserving theoretical guarantees.
    
    The work that is most related to ours is \cite{lagree2018algorithms}. Placed in a similar setting, it focuses on Online Influence Maximization with Persistence (OIMP). \cite{lagree2018algorithms} has a similar objective formulation, and proposes an algorithm called GT-UCB (for Good-Turing Upper Confidence Bound). The approach is inspired by the work of \cite{DBLP:journals/jmlr/BubeckEG13}, which used the Good-Turing estimator in a setting where a learning agent needs to sequentially select experts that only sample one of their potential elements at each round. Similar to rotting bandits, an adaptation of GT-UCB (called FAT-GT-UCB) is considered for scenarios where influencers may experience fatigue, i.e., a diminishing tendency to activate their user base as they are re-seeded during a campaign.  The key aspect that distinguishes our study from the one of \cite{lagree2018algorithms} is that \emph{we assume contextual information is known and exploitable in the sequential learning process}, as features of the influencers or of the information being diffused. In doing so, we provide solutions that are no longer agnostic to the information being diffused nor to the profiles of influencers, as was the case in \cite{lagree2018algorithms}. The contextual assumption leads to entirely different theoretical and algorithmic constructions, and is supported by our empirical evaluation. FAT-GT-UCB is one of our experimental baselines.

    Finally, we stress that in our bandit approach the parameters to be estimated throughout a campaign must capture how good an influencer still is (its \emph{remaining potential}). Hence a key difference with other multi-armed bandit studies for IM (\cite{DBLP:conf/kdd/WuLWCW19,NIPS2017_7137debd,DBLP:journals/jmlr/ChenWYW16, DBLP:conf/icml/VaswaniKWGLS17}) is that they look for a constant optimal seed set, while in our setting a round’s best action (choice of seeds) depends on the number of previous rounds and on the context.

\section{Problem Statement}\label{sec:problem}

\begin{table}[t]
	\vspace{-2mm}
	\caption{Summary of notations.}\label{tab:notations}
	\vspace{-7mm}
	\begin{center}
		\resizebox{\columnwidth}{!}{%
			\begin{tabular}{ l l }
				\toprule
				$T$ & total number of rounds in a campaign \\
				$K$ & total number of available influencers \\
				$Y_t$ & the context in round $t$ \\
				$I_t$ & the set of $L$ influencers selected in round $t$ \\
				$A_k$ & set of basic nodes reachable by influencer $k$ \\
				$S(I_t, Y_t)$ & the spread given by the environment in round $t$ \\
				$p_{k,j}(t)$ & the probability of influencer $k$ to activate basic node $j$ in round $t$ \\
				$\theta_{k,j}$ & feature vector that explains the probability of influencer $k$ \\
				& to activate basic node $j$ in round's context $Y_t$ \\  
				$n_k(t)$ & the history of number of selections of influencer $k$ in round $t$ \\
				$p(j)$ & the basic node's $j$ intrinsic probability of activating itself \\
				$\alpha$ & the external factor function which adjusts the basic node's\\& activation probability; e.g. defined as in Equation \ref{eq:alpha}.\\
				$F_t$ & the set of IDs of the activated basic nodes at the end of round $t$ \\
				$r_t$ & the reward at the end of round $t$ \\
				$r'_k(t)$ & the reward for the external factor's linear regression problem \\
				$C_j(t)$ & the cumulative Poisson count of activations for node $j$ in round $t$ \\
				$\theta_k$ & the influencer $k$'s feature vector \\
				$\hat{\theta}_k(t)$ & the estimator of the influencer $k$'s feature vector in round $t$ \\
				$\lambda_j$ & the Poisson intensity of activations for basic node $j$ \\
				$\lambda_k$ & the Poisson intensity of activations due to influencer $k$ \\
				$R_k(t)$ & the influencer $k$'s remaining potential (i.e. the feasible reward) in round $t$ \\
				$G_k(t)$ & Good-Turing estimator of the remaining potential for influencer $k$ \\
				$V_k(t)$ & design matrix updated by the context vectors in rounds when influencer $k$  is played \\
				$s_k(t)$ & the rewards history factor for linear regression \\
				\blue{$\gamma$} & the regularization factor for linear regression\\ 
			$b_k(t)$ & the UCB computed for influencer $k$ in round $t$ \\
			\bottomrule
		\end{tabular}
	}
\end{center}
\vspace{-6mm}
\end{table}

We formalize the IM problem, set in a discrete-time campaign consisting of $T$ rounds, with $K$ influencers among which the algorithm chooses seeds at each round. 

We model each influencer $k$ as having access to $A_k$ basic nodes, each one being influenced by $k$ with a probability $p_{k,j}(t), \forall j \in [1,\dots,A_k]$. We assume that $p_{k,j}(t)$ depends on each basic node's inner probability $p(j)$ of activating itself, on some $d$-dimensional profile $\theta_{k,j}$, and on the round's context.  In each round, a $d$-dimensional context $Y_t\in[0,1]^d$ is provided by the environment, in a similar manner to the contextual multi-armed bandit setting~\cite{DBLP:journals/ftml/Slivkins19}.  Considering that in our setting the reward is the number of \emph{newly} activated nodes, we assume also the impact of the number of selections of the influencer up to round $t$, $n_k(t)$, on the probability $p_{k,j}(t)$. Therefore, the probability of a basic node $j$ to be influenced by influencer $k$ is well-approximated by a function $\alpha(\langle \theta_{k,j}, Y_t \rangle, n_k(t))$ applied as a modifier to the basic node's inner activation probability $p(j)$. The modifier $\alpha$ is a function of the relation between the influencer, the basic node, and the round's context. Formally, the problem we study in this paper is defined as follows:

\vspace{-1mm}
\begin{problem}\hspace{-1mm}[Contextual Influence Maximization]\label{prob:cip}
Given a set of influencers $[K]={1,\dots,K}$, a budget of $N$ rounds (or trials), and a number $1 \leq L \leq K$ of influencers to be activated at each round, the objective is to solve the following optimization problem:
\vspace{-1mm}
\begin{equation}
\argmax_{I_t \subseteq [K], |I_t|=L, \forall 1\leq t\leq N} \mathbb{E}|\bigcup_{1\leq t\leq N} S(I_t, Y_t)|,
\vspace{-1mm}
\end{equation}
where $S(I_t, Y_t)$ is the spread of the chosen set of influencers for round $t$, and the probability that influencer $k$ activates basic node $j$ depends on the round's context $Y_t$ and  the number of $k$'s selections $n_k(t)$:
\vspace{-1mm}
\begin{equation}p_{k,j}(t) = \alpha(\langle \theta_{k,j}, Y_t \rangle, n_k(t)) p(j).
\vspace{-1mm}
\end{equation}
\end{problem}

A similar variant of this problem, which does not use contexts, was proven to be \textsf{NP}-hard in \cite{lagree2018algorithms}, and this hardness result immediately transfers to our problem (e.g., with a constant context for all rounds).

We now formulate the problem in a contextual bandit setting.  We assume a semi-bandit feedback at the end of each round, denoted $F_t$,  consisting of the set of IDs of the activated basic nodes. The \emph{reward} for the round is the number of new activations:
\vspace{-1mm}
\begin{equation}
r_t=\sum_{j=1}^{\bigcup_{k \in I_t}A_k} \mathbb{I}\{C_j(t)>0\} - r_{t-1}; r_0=0,
\vspace{-1mm}
\end{equation}
where $C_j(t)=\sum_{s=1}^t \mathbb{I}\{j\in F_s\}$ denotes for each basic node the number of times it has been activated. 

Given that the reward in each round is the number of \emph{newly} activated basic nodes, Problem~\ref{prob:cip} exhibits a \emph{diminishing returns property}: for each influencer, the expected number of new basic nodes it can activate decreases with each of its selections.

For each basic node $j$, its cumulative count of activations $C_j(t)$ up to round $t$ is a random quantity depending on the node's probability $p_{k,j}(t)$ of being activated by the played influencer; these activation probabilities are assumed to be unknown. As estimating all user profiles $\theta_{k,j}$ is computationally expensive, our goal will be instead -- given that the objective is to select the best influencer(s) at each round -- to directly estimate the influencers' potential based on the context at each round, as proxy for the probabilities of individual nodes.

To achieve this, we propose two algorithms that both assume a generalization $\theta_k$ of the unknown parameters $\theta_{k,j}$, and two different assumptions on the distribution of new activations for each influencer. More precisely, we assume that activations follow either (i) a Poisson distribution, given that they are counts of nodes, or (ii) a log-normal distribution, assuming that the scales of the rewards are normally distributed (in line with observations on the distribution of real-world social phenomena~\cite{sala2011brief}). In Section~\ref{The algorithm GLM-GT-UCB} we present the UCB-based solution that uses the Poisson distribution assumption, and in Section~\ref{Log-normal distribution} we present the LinUCB-based solution that assumes a log-normal distribution.

\section{GLM-GT-UCB Algorithm}\label{The algorithm GLM-GT-UCB}

The main idea behind the \glmucb~algorithm is to estimate the potential of each influencer, at each round, by some proxy measure. Here, by an influencer's \emph{potential} we understand the number of nodes that it can \emph{still} activate  (i.e., the reward); more formally, each influencer's remaining potential of activating new basic nodes in round $t$ is:
\vspace{-1mm}
\begin{equation}R_k(t)=A_k - \sum_{j=1}^{A_k} \mathbb{I}\{C_j(t-1)>0\}
\vspace{-1mm}
\end{equation}
The stochasticity of $C_j(t-1)$ means that the remaining potential is a random variable too. While this has been analyzed in the non-contextual case~\cite{lagree2018algorithms}, the challenge here is to account for the contextual dimension.  The proxy  we choose is the Good-Turing estimator~\cite{good1953population}, estimating the proportion of unseen items in a random process as the fraction of items seen only once (\emph{hapaxes}). 

There are two main technical challenges to modeling the remaining potential using Good-Turing estimators: (i) we are counting only \emph{new} activations, so a \emph{fatigue} factor needs to be added to the estimator, and
(2) the contextual case forces us to make an assumption on the model -- in our case, we opted for a generalized linear model using a Poisson distribution.

\subsection{\hspace{-3.5mm}Good-Turing with Poisson and External Factor}

An influencer's remaining potential is an unknown random variable. The Good-Turing estimator~\cite{good1953population}, adjusted with a fatigue function, was shown to successfully model an influencer's fatigue \cite{lagree2018algorithms}. The fatigue function, non-increasing in the number of influencer's selections, does not explicitly model an influencer's potential w.r.t. the diffused content. We thus propose a Good-Turing estimator adjusted by a function of the diffused content. 

For each basic node $j$, \eat{we know from Section \ref{sec:problem} that} its activation probability $p_{k,j}(t)$ is a function of (a) the linear combination of the node's feature vector $\theta_{k,j}$ and the round's context, and (b) the number of influencer's selections $n_k(s)$. The assumption we make is that the underlying distribution of each node's cumulative count of activations $C_j(t)$ is Poisson with intensities $\lambda_j \sum_{s=1}^t \sum_{k \in I_s} \alpha(\langle \theta_{k,j}, Y_s \rangle$, $n_k(s))$. Our approach is then to assume that the underlying distribution for the entire remaining potential of an influencer is Poisson with intensities $\alpha(\langle \theta_k, Y_t \rangle, n_k(t))\lambda_k, k \in \{1,\dots,K\}$, where the individual user response probabilities are small: $\lambda_k \geq \sum_{j=1}^{A_k} \lambda_j \ll A_k$. 
 Recall the true feature vector $\theta_k$ is initially unknown \eat{to the learning agent}, so its estimation  becomes a sub-problem of our problem. The classical solution is to use the regularized least-squares estimator: 
 \vspace{-1mm}
\begin{equation}
\hat{\theta}_k(t) = \argmin_{\theta \in \mathbb{R}^d} \sum_{s=1}^{t-1} (r^{'}_k(s) - \langle \theta, Y_t \rangle)^2 + \gamma\|\theta\|_2^2),
\vspace{-1mm}
\end{equation} where $r^{'}_k(s)$ is the round's reward (adapted for the sub-problem) and $\gamma$ is the penalty factor that ensures the solution's uniqueness; more details are given in Sec. \ref{The external factor}.

After $t$ rounds, we observe the cumulative Poisson counts $C_j(t)$ of activations of each basic node $j \in \{1,\dots,A_k\}$ by the corresponding influencer. The cumulative counts are distributed with rate 
\vspace{-1mm}
\begin{equation}
\lambda_j \sum_{s=1}^t \sum_{k \in I_s} \alpha(\langle \theta_{k,j}, Y_s \rangle, n_k(s)),
\vspace{-1mm}
\end{equation}
and in estimation with rate
\vspace{-1mm}
\begin{equation}
\lambda_j \sum_{s=1}^t \sum_{k \in I_s} \alpha(\langle \theta_k, Y_s \rangle, n_k(s)).
\vspace{-1mm}
\end{equation}
Thus, the remaining potential can be expressed as the conditional expectation of cumulative counts of new basic nodes that would be influenced in round $t$:
\vspace{-1mm}
\begin{equation}
R_k(t) = \sum_{j=1}^{A_k}\lambda_j\alpha(\langle \theta_k, Y_t \rangle, n_k(t))\mathbb{I}\{C_j(t-1)=0\}
\vspace{-2mm}
\end{equation}
The expectation of $k$'s remaining potential in round $t$ is
\vspace{-4mm}
\begin{equation}
\mathbb{E}[R_k(t)] = \alpha(\langle \theta_k, Y_t \rangle, n_k(t)) \cdot 
\vspace{-5mm}
\end{equation}\label{eq:erm}
\[
\vspace{-1mm}
\sum_{j=1}^{A_k} \lambda_j e^{-\lambda_j \sum_{s=1}^{t-1} \sum_{k^{'} \in I_s} \alpha(\langle \theta_{k^{'}}, Y_s \rangle, n_{k^{'}}(s))}
\]
\glmucb~estimates $k$'s remaining potential by:
\vspace{-1mm}
\begin{equation}
G_k(t) = \alpha(\langle \hat{\theta}_k(t), Y_t \rangle, n_k(t)) \frac{1}{n_k(t)} \sum_{j=1}^{A_k} \sum_{s=1}^{t-1} \cdot 
\vspace{-2mm}
\end{equation}
\[
\hspace{-0.2cm}
\frac{\mathbb{I} \{X_{s,j,k}=1, \{X_{s,j, k^{'}}=0\}_{k^{'} \in I_s \setminus \{k\}}, \{X_{l,j, k^{'}}=0\}_{l \neq s, k^{'} \in I_l}\}}{\alpha(\langle \hat{\theta}_k(s), Y_s \rangle, n_k(s))}
\vspace{-1mm}
\]
where $n_k(t)$ is the number of selections of influencer $k$ up to round $t$, $X_{s,j,k}$ is a binary random variable equal to 1 when $j$ is activated in round $s$ by influencer $k$, $l \in \{1, 2,\dots, t-1\}, k^{'} \in I_l \subseteq[K]$. We discuss next the external factor estimated through regular linear regression, used to regulate the proportion of hapaxes in the cascades generated by influencer $k$.

\vspace{-.1cm}
\subsection{The External Factor $\alpha$.}\label{The external factor}

The external factor, as stated before, is a sub-problem of Problem~\ref{prob:cip}. The remaining potential of an influencer is modelled by the combination of the external factor and the average count of hapaxes from the Good-Turing estimator. Under the assumption of a Poisson distribution for the rewards, and their property of diminishing returns, the external factor can be chosen as an adaptation of the inverse link function (mean function) for the Poisson distribution: 
\begin{equation} \label{eq:alpha}
\alpha(\langle \theta_k, Y_t \rangle, n_k(t)) =  e^{f(n_k(t))\left(\langle \theta_k, Y_t \rangle\right)}
\end{equation} 
 The $f(n_k(t))$ function is assumed to be non-increasing, models the influencer's fatigue, and depends on the influencer's selections. 
By combining the two estimators, the predicted values for this sub-problem are:
\vspace{-1mm}
 \begin{equation} \label{eq:externalFactorReward}
r'_k(t) = \frac{\ln\left(\frac{r_t n_k(t)}{\sum_{j=1}^{A_k} \sum_{s=1}^{t-1}  \frac{\text{hapax}_{s,j,k} }{\alpha(\langle \hat{\theta}_k(s), Y_s \rangle, n_k(s))}}\right)}{f(n_k(t))}, \text{where}
\vspace{-3mm}
 \end{equation}
\vspace{-2mm}
\begin{equation}
\text{hapax}_{s,j,k} =
\vspace{-1mm}
\end{equation}
\[
\mathbb{I} \{X_{s,j,k}=1, \{X_{s,j, k^{'}}=0\}_{k^{'} \in I_s \setminus {k}}, \{X_{l,j, k^{'}}=0\}_{l \neq s, k^{'} \in I_l}\}
\]

The argument of the external factor function is a random variable $r'_k(t) =  \langle \theta_k, Y_t \rangle + \eta_t$. The noise $\eta_t$ is assumed conditionally 1-subgaussian. The regularized least-squares estimator for the feature vector is:
\vspace{-2mm}
\begin{equation}
\hat{\theta}_k(t) = V_k^{-1}(t) \sum_{s=1}^{t-1} Y_s r^{'}_k(s) \mathbb{I}\{k \in I_s\},
\vspace{-1mm}
\end{equation}
where $V_k(t) = \gamma\textit{I} + \sum_{s=1}^{t-1} Y_s Y_s^T \mathbb{I}\{k \in I_s\}$; $\gamma \ge 0$ is the penalty factor that ensures an unique solution. The design matrix $V_k(t)$ is computed from the contexts of the rounds in which the corresponding influencer was played, adjusted by its number of the selections.

\subsection{Upper-Confidence Bound.}\label{The upper confidence bound}

UCB algorithms provide a disciplined balance between the exploitation of the options that are known as best up to the decision round, and the exploration of the ones for which the learning agent has not acquired enough information yet. The \glmucb~ algorithm follows the main lines of an UCB-based algorithm, and its flow is presented in Algorithm~\ref{GLM-GT-UCB}. It starts with an initialization phase, where each influencer is played once in a random context. The observed rewards are used to initialize the influencer's statistics, necessary for further decisions. 
For the Good-Turing estimator, we maintain the number of selections $n_k(t)$ and the history of the discounted rewards for computing this estimator, as well as the sample-mean activations for computing the UCB index. For the linear regression of the external factor, we maintain a history of the rewards and the design matrix for each influencer:
\vspace{-4mm}
\begin{equation}V_k(t) = \gamma \textit{I}_d + \sum_{s=1}^{t-1} Y_s Y_s^T \mathbb{I}\{k \in I_s\}
\vspace{-5mm}
\end{equation}
\vspace{-2mm}
\begin{equation}s_k(t) =  \sum_{s=1}^{t-1} Y_s r'_k(s) \mathbb{I}\{k \in I_s\}.
\vspace{-1mm}
\end{equation}
To make better use of the available contextual information, the contextual UCB can be added to the estimated external factor:
\begin{equation} \hspace{-3.3mm} \label{eq:alpha}
\alpha(\langle \hat{\theta}_k, Y_t \rangle, n_k(t))\hspace{-0.75mm}=\hspace{-0.75mm}e^{f(n_k(t))\left(\langle \hat{\theta}_k, Y_t \rangle + \gamma \sqrt{Y_t^T V_k^{-1}(t) Y_t}\right)}
\end{equation}

In each subsequent round, the agent gets the context from the environment. It estimates for each influencer its feature vector, by the regularized least-square estimator in the stochastic linear bandit $\hat{\theta}_k(t)$, which is then used to compute the estimator of the remaining potential. The UCB $b_k(t)$ is obtained by adding the confidence factor $\beta_k(t)$. The agent plays the influencers with the highest UCBs, observes and divides the reward equally among them, and updates their statistics.

The UCB index computed on the adapted Good-Turing estimator captures both the confidence in the unmodified Good-Turing estimator, and the one in the estimator of the influencer's true unknown vector $\theta_k$:
\vspace{-1mm}
\begin{multline}\label{equ:bkt_glm_gt_ucb}
b_k(t) = G_k(t) + \beta_k(t)
+ \\
 \hat{\lambda}_k(t) \left( 1 - e^{\frac{-2+ C_k(t)}{n_k(t)}} \frac{1}{n_k(t)} \sum_{s=1}^{t-1} e^{\frac{-2- C_k(s)}{n_k(s)}} \right) \text{, where}
\end{multline}
\begin{multline}\label{equ:bkt}
\beta_k(t) = 
\sqrt{\frac{2\hat{\lambda}_k(t) e^{\frac{3+ 2C_k(t)}{n_k(t)}} \sum_{s=1}^{t-1} e^{ \frac{2 - 2C_k(s)}{n_k(s)}}}{n^2_k(t)}\ln \frac{1}{\delta}}
 + \\
\sqrt{\frac{ e^{2/n_k(t)} \hat{\lambda}_k(t) \ln(1/ \delta)}{\sum_{s=1}^{t-1} \sum_{k^{'} \in I_s} e^{-1/n_{k^{'}}(s)}}} +
 \frac{e^{ \frac{1+ C_k(t)}{n_k(t)}}\sum_{s=1}^{t-1}e^{ \frac{1+ C_k(s)}{n_k(s)}}}{3n_k(t)} \ln\frac{1}{\delta} ,
\end{multline}
$\hat{\lambda}_k(t) = \frac{1}{n_k(t)}\sum_{s=1}^{t-1} \frac{|F_s|}{L}\mathbb{I}\{k \in I_s\}$ is the sample-mean number of activations for influencer $k$, and $C_k(t) = \gamma \|Y_t\|_{V_k^{-1}(t)}$ is the contextual UCB for the external factor.

\begin{algorithm}[t]
	\small
	\caption{\glmucb}\label{GLM-GT-UCB}
	\begin{algorithmic}[1]
		\State \textbf{Input:} influencers K, rounds budget T,  external factor function $\alpha$, regularization factor $\gamma$,  fatigue function $f$, number of selections $L$
		\State \textbf{Initialization:} play each influencer $k \in [K]$ once in given random contexts $Y_t$, observe the reward $r_t, t\in[K]$, and update the statistics $n_k(1) = 1, \hat{\lambda}_k(1) = |F_k|$ for the Good-Turing estimator, and $V_k(1)=\gamma\textit{I} + Y_t Y_t^T, s_k(1) = Y_t r'_k(1)$~\footnotemark ~ for the external factor. 
		\For{ $t=K + 1,\dots,T$}
		\State Get the context $Y_t$
		\For{ $k \in [K]$ }
		\State Estimate the unknown vector:
		\begin{equation}\hat{\theta}_k(t) = V^{-1}_k(t)s_k(t)\end{equation}
		\State \hspace{-1mm}Compute UCB for remaining  potential estimator
		\vspace{-1mm}
		\begin{equation}b_k(t) = G_k(t) + \beta_k(t),
			\vspace{-1mm}\end{equation}
		\vspace{-3mm}
		\begin{equation}G_k(t) = \alpha(\langle \hat{\theta}_k(t), Y_t\rangle, n_k(t)) \cdot 
			\vspace{-1mm}\end{equation}
		\[ \frac{1}{n_k(t)} \sum_{j=1}^{A_k}\sum_{s=1}^{t-1} \frac{\text{hapax}_{s,j,k}}{\alpha(\langle \hat{\theta}_k(s), Y_s \rangle, n_k(s))} \]
		~~~~~~~~  and $\beta_k(t)$ is given by the confidence interval.
		\EndFor \State \textbf{end for}
		\State Choose set $I_t$ of $L$ influencers with largest UCB
		\State Play 
		the chosen influencers, observe spread, divide it equally among influencers, update their statistics:
		\For{$k^{'} \in I_t$}
		\State Update $r_{k^{'}}(t)$ by Eq. (\ref{eq:externalFactorReward}).; 
		$ n_{k^{'}}(t+1)=n_{k^{'}}(t)+1;
		V_{k^{'}}(t+1) = V_{k^{'}}(t) + Y_t Y_t^T; s_{k^{'}}(t+1) = s_{k^{'}}(t) + Y_t r'_{k^{'}}(t)$
		\EndFor \State \textbf{end for}
		\EndFor \State \textbf{end for}
	\end{algorithmic}
\end{algorithm}

\footnotetext{$r'_k(1)=\frac{1}{f(1)}\ln\left(\frac{r_t}{\sum_{j=1}^{A_k} \sum_{s=1}^{t-1} \frac{ \text{hapax}_{s,j,k}}{\alpha(\langle \hat{\theta}_k(s), Y_s \rangle, 1)}}\right). $}

\vspace{-1mm}
\subsection{Theoretical Analysis}
The UCB index is chosen as the maximum difference that can occur between the GT estimator and the true remaining potential with some chosen confidence. Theorem~\ref{thm:ucb_confidence} provides the confidence interval for the estimated remaining potential. Its proof has three steps: the concentration of the true remaining potential, the concentration of the Good-Turing estimator,  the bias of the estimator.
\vspace{-2mm}
\begin{theorem}\label{thm:ucb_confidence}
With probability at least $1-\delta$, having the expected activations $\lambda_k=\sum_{j=1}^{A_k} p_{k,j}(t)$, and $\beta_k(t)$ set as in Equation \ref{equ:bkt}
, we have
\vspace{-2mm}
\begin{multline}
\hspace{-2mm}
-\beta_k(t) + 
\Omega\left(\frac{T\lambda_k(T)}{n_k(T)}e^{\frac{C_k(T)}{n_k(T)}}\right)
\leq 
\vspace{-2mm}
\\
\hspace{-2mm}
R_k(t) - G_k(t) \leq \beta_k(t) +
\mathcal{O}\left(T\frac{\lambda_k(T)}{n_k(T)}e^{\frac{C_k(T)}{n_k(T)}}\right)
\end{multline}
\end{theorem}
\vspace{-2mm}

\vspace{-3mm}

\section{Log-normal Distribution}\label{Log-normal distribution}
We now consider the second alternative, that the underlying distribution is a log-normal one. Now, the influence maximization problem can be solved by an adapted LinUCB \cite{chu2011contextual}. LinUCB computes the expected reward of each arm by finding a linear combination of the previous rewards of the arm. It estimates the unknown parameter $\theta_t$ of the current round as a linear combination of the previously seen feature vectors and rewards, and it estimates the expected reward on the current round by linearly combining it with the current feature vector. The adaptation of LinUCB to our problem consists in maintaining a design matrix per influencer, which is updated by the context of the round in which the influencer has been played. This change implies that a separate parameter is estimated for each influencer, and its linear combination with the current round's context will estimate the reward at logarithmic scale. Note that the linear combination estimates the scale of the reward since we assume that the rewards are log-normally distributed. The main flow is presented in Algorithm~\ref{Lognorm LinUCB}. It is similar to the one of LinUCB~\cite{chu2011contextual}, in that at each step it chooses the best reward in terms of the linear combination of the context and the learned profile plus a confidence bound. In general linear models -- of which \linucb\ is part of -- this bound is based on a design matrix $V_k$ and the given context $Y_t$.

\vspace{-2mm}
\subsection{Regret analysis.} The regret analysis is performed at logarithmic scale; this restriction stems from having the logarithm of the new activations being normally distributed. In \cite{chu2011contextual},  theoretical guarantees for LinUCB were challenging, due to the lack of independence of the random variables for the rounds' rewards.  The solution was to use a supporting algorithm, SupLinUCB, estimating the unknown parameter only from the feature vectors and rewards from the rounds in which the agent performs random exploration. Each round is split into levels, and each level maintains an index set used for learning, comprising the indices of the rounds with independent rewards. When exploring, the round is added to the index set of the corresponding level. \eat{in which the choice is made.}

\begin{algorithm}[t]
\small
\caption{\linucb}\label{Lognorm LinUCB}
\begin{algorithmic}[1]
\State \textbf{Input:} influencers $K$, selections $L$, $\gamma \in \mathbb{R}_+$,  $d \in \mathbb{N}$
\State $V_k(1) = I_d, \forall k \in [K]$ and  $s_k(1) = \mathbf{0}_d, \forall k \in [K]$
\For{t=1,\dots,T}
    \State Get context $Y_t$
    \For{$k \in [K]$}
        \State $\hat{\theta}_k(t) = V^{-1}_k(t) s_k(t)$
        \State $b_k(t) = \langle \hat{\theta}_k(t), Y_t \rangle + \gamma \sqrt{Y_t^T V^{-1}_k(t) Y_t}$
    \EndFor \State \textbf{end for}
    \State Choose set $I_t$ of $L$ influencers with largest UCB $b_k(t)$
    \State Observe spread, compute reward $r$ by discounting previously activated basic nodes and dividing by $L$.
    \For{$k^{'} \in [I_t]$}
        \State $V_{k^{'}(t)}(t+1) = V_{k^{'}(t)}(t) + Y_t Y^T_t$ 
        \State $s_{k^{'}(t)}(t+1) = s_{k^{'}(t)}(t) + \ln(r) Y_t$
    \EndFor \State \textbf{end for}
\EndFor \State \textbf{end for}
\end{algorithmic}

\vspace{-1mm}

\end{algorithm}

We designed similarly IM-SupLinUCB and its sub-routine IM-Base\-LinUCB, preserving the steps of SupLinUCB \cite{chu2011contextual} and SupLinRel \cite{auer2002using}. For each influencer, the UCB is computed for the scale of the reward -- the new activations. 

\noindent IM-BaseLinUCB and IM-SupLinUCB's analysis is skipped here, as it is  similar to \cite{chu2011contextual,auer2002using}. The regret for stochastic linear bandits is generally defined as:
\vspace{-2mm}
\begin{equation}
    \mathcal{\hat{R}}_t = \sum_{s=1}^t \max_{k \in [K]} \langle \theta_k, Y_t\rangle - \sum_{s=1}^t r_s
    \vspace{-5mm}
\end{equation}
    \vspace{-1mm}
\begin{equation}
    \mathcal{R}_t = \mathbb{E}[\mathcal{\hat{R}}_t] = \mathbb{E}\left[ \sum_{s=1}^t \max_{k \in [K]} \langle \theta_k, Y_t\rangle - \sum_{s=1}^t r_s\right]
        \vspace{-1mm}
\end{equation}

We have the following $\tilde{\mathcal{O}}(\sqrt{T})$ regret bound for the supporting algorithm on logarithms of rewards:

\begin{theorem}\label{thm:linucb_regret}
If IM-SupLinUCB uses parameter $\gamma = \sqrt{\frac{1}{2}\ln(\frac{2TK}{\delta})}$, with probability $1-\delta$ the regret of \linucb~at logarithmic scale is
\begin{equation}\hspace{-2mm}\mathcal{\hat{R}}_t \leq 2\sqrt{T} + 44K\left(1+\ln\left(2TK\ln(T)/\delta\right)/2\right)^{\frac{3}{2}}\sqrt{Td}
\vspace{-1mm}
\end{equation}
\end{theorem}
\begin{proof}
Proof  similar to that of  \cite[Theorem 6.]{auer2002using}.
\end{proof}

\vspace{-3mm}
\section{Experiments}

 \begin{figure*}[h]
  \vspace{-4mm}
  \centering
  \includegraphics[width=1\linewidth]{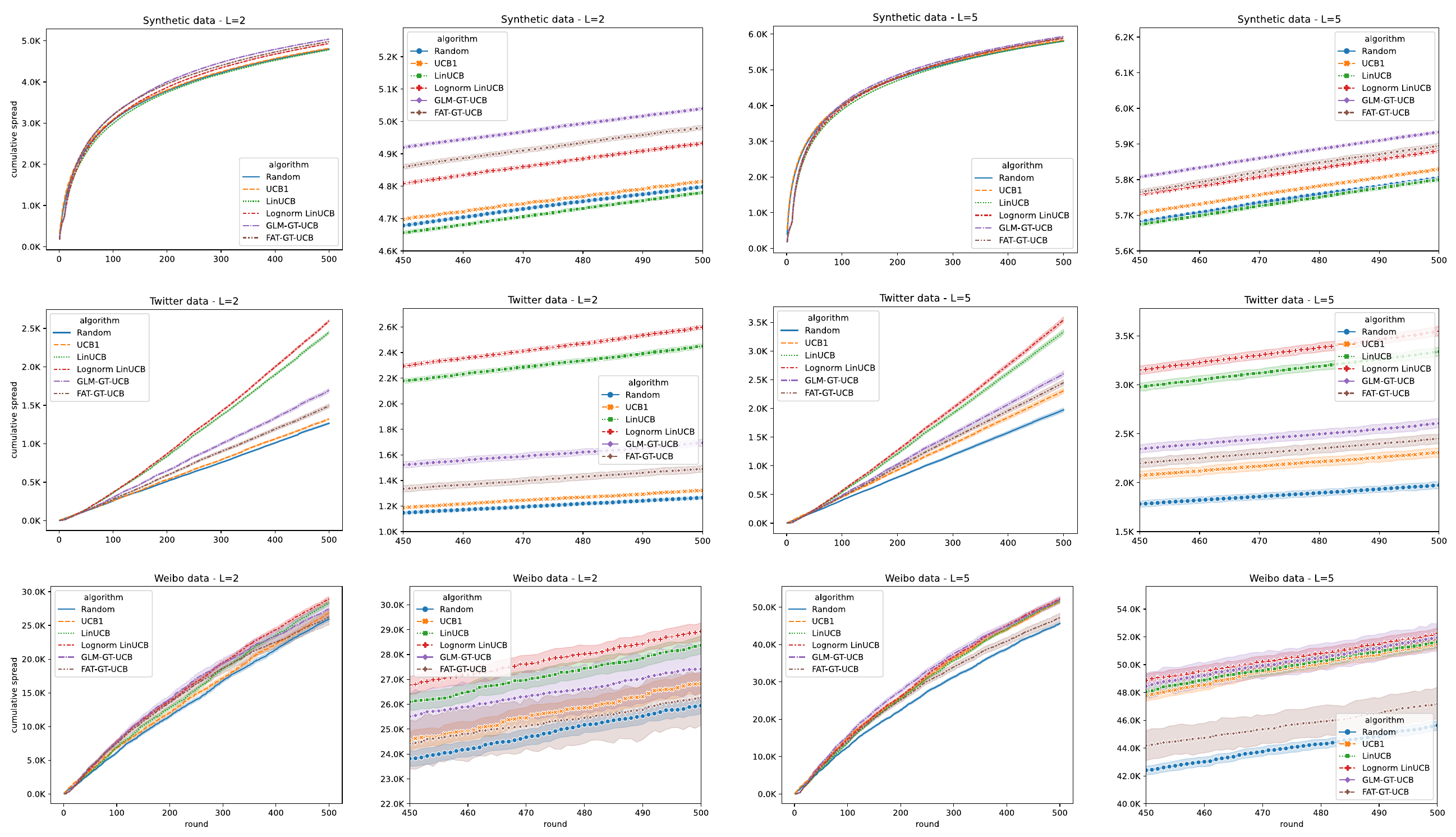}
  \vspace{-1.5mm}
\caption{\hspace{-1mm}Cumulative rewards -- L=2, L=5 -- normal plot (odd rows) \& plot zoomed to last 50 rounds (even rows).}
\vspace{-5.5mm}
\label{fig:results}
\end{figure*} 

We tested \glmucb~and \linucb~on synthetically generated data, on data we collected from Twitter, and on a publicly available dataset from Sina Weibo~\cite{zhang2013social}. All the results are averaged over $100$ runs.

\textit{Empirical distribution of rewards.}
First, as additional support for assuming in \glmucb\ a Poisson distribution for each influencer's rewards, we inspected in Twitter the reward distributions per influencer and context. As the reward is a dynamic quantity -- depends on previously activated nodes -- we generated random campaigns, and used an \textsc{Oracle} to select the influencer with the highest reward at each round. 
We plotted for each influencer and context the empirical distribution of its rewards and the Poisson fit, which supports the assumption for the underlying distributions. Fig. \ref{fig:distribution} of the supplementary material illustrates such a distribution, the one with the largest number of samples. \eat{Note that we cannot perform a similar test in Weibo, due to the large dimensionality of contexts (100 topics), and thus higher contextual diversity, leading to too few samples to create an empirical distribution of rewards for an influencer in a given context.}

\textit{Synthetic data experiments.}

The synthetic data is generated starting from the premise that each basic node's activation probability is known. Therefore, all the edges and nodes are assumed to be known as well. The synthetic graph is randomly generated following the Barab\'{a}si-Albert preferential attachment model~\cite{barabasi2016network}. The model's parameters are chosen as follows: $30{,}000$ nodes and, at each step, one new edge to be attached from new nodes to existing ones. Then, the $10$ nodes having the maximum degrees are chosen to be the influencers.

Activation probabilities are computed as a sigmoid function of the inner product of the context and the basic node's feature vector plus some random small noise.
This is preferable in order to project the results intro probability thresholds, i.e., the value over which the node is considered activated - $0.999$ in our experiments. The inner product captures the linear relationship between context and hidden profile. For each node, its feature vector is randomly generated from a normal distribution. Then, the context of a campaign's round  is generated from another normal distribution. A given round is chosen to be viral with a $50\%$ probability, i.e., the distribution from which the context is drawn is chosen such that its inner product with most of the basic user feature vectors results in higher values for the activation function. For these rounds, only $L+1$ influencers are chosen to use the viral context. The diffusion model is assumed to be IC \cite{kempe2003maximizing}, a campaign consist of $500$ rounds, and  results are averaged over $100$ runs. \eat{The \emph{activation probability threshold}, i.e., the value of the sigmoid function over which the node is considered to be activated, is set to $0.999$.} $\gamma$ is set to $\sqrt{1/2\ln{(\sqrt{2TK/\delta)}}}$ everywhere.

\textit{Baselines.} We compare against Random,
UCB1~\cite{auer2002finite}, LinUCB~\cite{chu2011contextual}, and FAT-GT-UCB~\cite{lagree2018algorithms}. The random policy chooses a random influencer in each round. UCB1 is a well-known algorithm in the bandit literature, one which does not model contexts. The FAT-GT-UCB algorithm models the influencer's fatigue in a context-free setting.
The results (Fig.~\ref{fig:results}, top row) show that \glmucb~and \linucb~are both capable of learning the remaining potential of influencers from their performance in different contexts. Making decisions based on the available information about the round's context has a clear added value, compared to only considering the time-based fatigue of approaches such as FAT-GT-UCB.

\textit{Twitter dataset.} We extracted from Twitter logs a collection of retweets.  These can be viewed as belonging to basic nodes, representing successful activations of the original tweets from influencers. To test the capability of the algorithms to choose the right influencers for a given context, we extracted from tweet the round's context.  As in \cite{DBLP:conf/cikm/ShinCLD15}, a tweet is encoded into a multi-dimensional vector. The encoding represents the distribution of the tweets' words over a predefined number of centroids (24 in our experiments). The centroids are obtained via clustering ($K$-means) on the public vocabulary \emph{glove-twitter-200}\footnote{\url{https://nlp.stanford.edu/projects/glove/}} from the word embedding open-source library Gensim\footnote{\url{https://github.com/RaRe-Technologies/gensim-data}}. Each word is assigned to its closest centroid, thus obtaining the distribution. The largest cluster is split into 5 smaller clusters.

In Twitter and Weibo, we improve the learning rate of ~\glmucb~ by adding $10/L$ activations only when learning the external factor via linear regression. The plotted results are with the true values of activations.

The campaigns are created by randomly choosing the context for each round to be one of the available centroid distributions in the dataset. We chose the set of influencers to be the users with the highest degrees. In each round, each algorithm chooses which influencers it wants to play. Due to the sparsity in the data, we implemented the bandit to sample with replacement from the set of all tweets with the round's context matching their centroid distribution and the algorithm's chosen influencer as the original user id. If there is no log for this tuple, we consider that no basic node has been activated. The reward is computed by discounting previously activated basic nodes. 

The results are in Fig.~\ref{fig:results} (middle row), for either the entire campaign of $500$ rounds or zoomed on rounds $450$ to $500$; the shaded areas represent the uncertainty.

\textit{Weibo dataset.} Using a public dataset from the popular Chinese microblogging platform, 
we designed the experiment as in the Twitter scenario. The topic distributions created by \cite{zhang2013social} are used as contexts. There are 100 topics, and for each post the distribution of topics is computed by using Latent Dirichlet Allocation \cite{heinrich2005parameter}. Once again, in Fig. ~\ref{fig:results} (bottom row) we can see that our methods manage to perform better by using the round's context information when selecting influencers. The relative performance can depend on time: \glmucb~ seems to initially learn faster.

 From both experiments on real-world datasets, we can conclude that our approaches -- especially \linucb\ -- are capable of learning viral cascades in different datasets and cascade settings, which increases their potential in  spread maximization (visible in the ``steps'' of the plots); this is not the case with other approaches, which seem to work best when the cascades have fewer outliers in terms of size; hence, they do not learn quickly enough to adapt.


\section{Conclusion}
We presented in this paper the problem of designing advertising campaigns from the point of view of contextual influence maximization, when the exact diffusion model is not fully exploitable. By adapting approaches from the contextual bandit literature, we designed algorithms \glmucb\ and \linucb, using different assumptions on the underlying distributions of the number of influenced nodes: Poisson  and log-normal respectively. We showed both theoretically and experimentally that our approaches have the potential to learn the influencers' potential, leading to improved IM campaigns compared to other state-of-the-art methods.


\balance

\section*{Acknowledgments}
We thank Olivier Cappé and Yoan Russac, for early discussions and ideas on modeling the distribution of rewards. This work was also supported by the Singapore NRF DesCartes research grant.

\bibliographystyle{plain}
\vspace{-3mm}
\bibliography{bibliography}


\clearpage
\appendix

\section{Other empirical results}

\begin{figure}[t]
\includegraphics[width=1\linewidth]{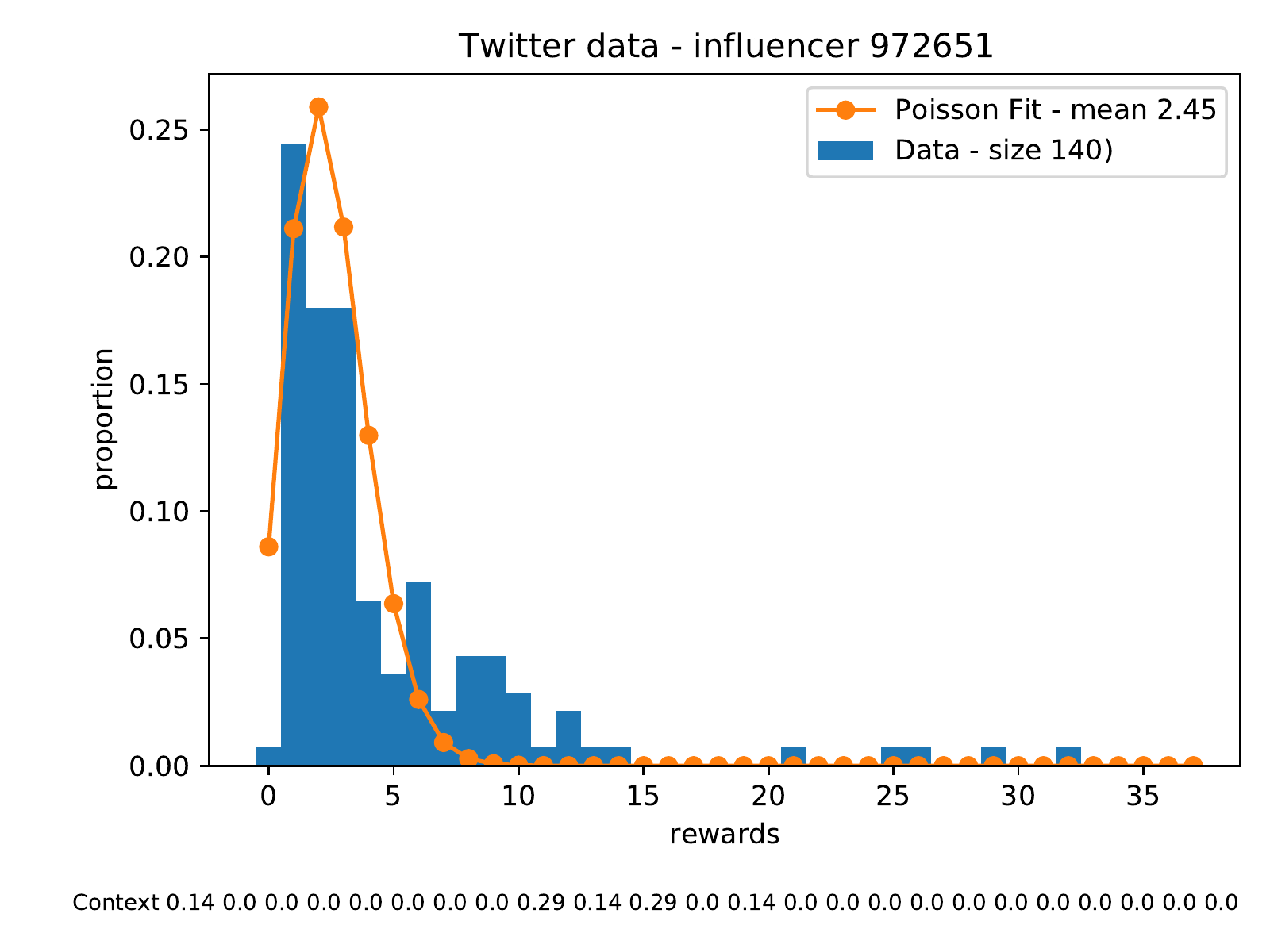}
\caption{Reward distributions in Twitter.}
\label{fig:distribution}
\end{figure}

As additional empirically support for assuming in \glmucb\ a Poisson distribution for each influencer's rewards, we inspected the reward distributions per influencer and context, in the Twitter dataset. Given that the reward is a dynamic quantity -- as it depends on previously activated nodes -- we generate random campaigns, and use an \textsc{Oracle} to select the influencer with the highest reward for each round. 
Then we plot for each influencer and context the empirical distribution of its rewards and the Poisson fit. The results support the assumption for the underlying distributions. The Fig. \ref{fig:distribution} (supplementary material) illustrates the distributions with the largest number of samples. Note that we cannot perform a similar test in Weibo, due to the large dimensionality of contexts (100 topics), and thus higher contextual diversity, leading to too few samples to create an empirical distribution of rewards for an influencer in a given context.

\section{\glmucb~ - Theoretical analysis}\label{demglmucb}

\begin{table}[h!]
\begin{center}
\caption{Assumptions}
\begin{tabular}{ l }
 \hline
 $\theta_{k,j} \approx \theta_k$ (the influencer's unkown vector $\theta_k$ captures  \\
 the basic node's trend in activating in a context\\
 along its inner probability $p(j)$)  \\
 $\alpha(\langle\theta_k, Y_t\rangle,n_k(t)) = e^{f(n_k(t))\langle\theta_k, Y_t\rangle}$\\
 $f(n_k(t)) = \frac{1}{n_k(t)}$ \\
 $\|\theta_k\|\leq 1$ \\
 $\|Y_t\| \leq 1$ \\
 \hline
\end{tabular}
\end{center}
\end{table}

\begin{proof}

The theoretical analysis is performed for one influencer $k$; it follows the same steps for any other influencer.

\textbf{Preliminaries:}

\begin{equation}
\begin{split}
    p_{k,j}(t) & = \alpha(\langle\theta_k, Y_t\rangle,n_k(t)) p(j)
\end{split}
\end{equation}

\begin{multline}
    P(C_j(t-1) = 0) =  
    e^{-\lambda_j \sum_{s=1}^{t-1} \sum_{k^{'} \in I_s} \alpha(\langle\theta_{k^{'}}, Y_s\rangle, n_{k^{'}}(s))}
\end{multline}

\begin{equation}
\begin{split}
    P(C_{k,j}(t-1) = 1) & =
    \sum_{s=1}^{t-1}
    \lambda_j \alpha(\langle \theta_k, Y_s \rangle, n_k(s)) 
    \cdot \\ &
    e^{-\lambda_j \sum_{l=1}^{t-1} \sum_{k^{'} \in I_l} \alpha(\langle\theta_{k^{'}}, Y_l\rangle, n_{k^{'}}(l))}
\end{split}
\end{equation} 

\begin{multline}
    R_k(t) = \alpha(\langle\theta_k, Y_t\rangle,n_k(t))\sum_{j=1}^{A_k}\lambda_j \mathbb{I}\{C_j(t-1)=0\}
\end{multline}

\begin{multline}
    \mathbb{E}[R_k(t)] = \alpha(\langle\theta_k, Y_t\rangle,n_k(t))
    \cdot \\
    \sum_{j=1}^{A_k}\lambda_j e^{-\lambda_j \sum_{s=1}^{t-1} \sum_{k^{'} \in I_s} \alpha(\langle\theta_{k^{'}}, Y_s\rangle, n_{k^{'}}(s))}
\end{multline}

\begin{multline}
    G_k(t) = \alpha(\langle\hat{\theta}_k(t), Y_t\rangle,n_k(t)) \frac{1}{n_k(t)} \sum_{j=1}^{A_k} \sum_{s=1}^{t-1} 
    \cdot \\
    \frac{\mathbb{I}\{X_{s,j,k}=1,\{X_{l,j,k^{'}}=0\}_{k^{'}\in I_s\setminus \{k\}}, \{X_{l,j,k^{'}}=0\}_{l\neq s, k^{'}\in I_l}\} }{\alpha(\langle\hat{\theta}_k(s), Y_s\rangle, n_k(s))}
\end{multline}

\begin{multline}
    \mathbb{E}[G_k(t)] = \alpha(\langle\hat{\theta}_k(t), Y_t\rangle,n_k(t))\frac{1}{n_k(t)}
    \sum_{j=1}^{A_k} \sum_{s=1}^{t-1}  
    \cdot \\
    \frac{\alpha(\langle\theta_k, Y_s\rangle, n_k(s))\lambda_j e^{-\lambda_j\sum_{l=1}^{t-1} \sum_{k^{'}\in I_l}\alpha(\langle\theta_{k^{'}}, Y_l\rangle, n_{k^{'}}(l))}}{\alpha(\langle\hat{\theta}_k(s), Y_s\rangle, n_k(s))}
    \\
    =\frac{\alpha(\langle\hat{\theta}_k(t), Y_t\rangle,n_k(t))}{\alpha(\langle\theta_k, Y_t\rangle,n_k(t))} \frac{1}{n_k(t)} \sum_{j=1}^{A_k}\alpha(\langle\theta_k, Y_t\rangle,n_k(t))\lambda_j
    \cdot \\
     e^{-\lambda_j\sum_{l=1}^{t-1}\sum_{k^{'}\in I_l}\alpha(\langle\theta_{k^{'}}, Y_l\rangle, n_{k^{'}}(l))}
    \sum_{s=1}^{t-1}  \frac{\alpha(\langle\theta_k, Y_s\rangle, n_k(s))}{\alpha(\langle\hat{\theta}_k(s), Y_s\rangle, n_k(s))}
    \\
    =\frac{\alpha(\langle\hat{\theta}_k(t), Y_t\rangle,n_k(t))}{\alpha(\langle\theta_k, Y_t\rangle,n_k(t))}
    \frac{1}{n_k(t)} \sum_{s=1}^{t-1}  \frac{\alpha(\langle\theta_k, Y_s\rangle, n_k(s))}{\alpha(\langle\hat{\theta}_k(s), Y_s\rangle, n_k(s))}
    \cdot \\
    \alpha(\langle\theta_k, Y_t\rangle,n_k(t))\sum_{j=1}^{A_k}\lambda_j e^{-\lambda_j\sum_{l=1}^{t-1}\sum_{k^{'}\in I_l}\alpha(\langle\theta_{k^{'}}, Y_l\rangle, n_{k^{'}}(l))}
    \\
    =\frac{\alpha(\langle\hat{\theta}_k(t), Y_t\rangle,n_k(t))}{\alpha(\langle\theta_k, Y_t\rangle,n_k(t))} \frac{\mathbb{E}[R_k(t)]}{n_k(t)} 
    \sum_{s=1}^{t-1} \frac{\alpha(\langle\theta_k, Y_s\rangle, n_k(s))}{\alpha(\langle\hat{\theta}_k(s), Y_s\rangle, n_k(s))}
\end{multline}

\textbf{The bias of the remaining potential estimator:}

\begin{multline}
    \mathbb{E}[R_k(t)] - \mathbb{E}[G_k(t)] = \mathbb{E}[R_k(t)] -
    \cdot \\
    \frac{\mathbb{E}[R_k(t)]}{n_k(t)}
    \frac{\alpha(\langle\hat{\theta}_k(t), Y_t\rangle,n_k(t))}{\alpha(\langle\theta_k, Y_t\rangle,n_k(t))}
    \sum_{s=1}^{t-1}  \frac{\alpha(\langle\theta_k, Y_s\rangle, n_k(s))}{\alpha(\langle\hat{\theta}_k(s), Y_s\rangle, n_k(s))}
\end{multline}

\begin{multline}
    \mathbb{E}[R_k(t)] - \mathbb{E}[G_k(t)] = \mathbb{E}[R_k(t)] 
    ( 1 - 
    \\
    \frac{1}{n_k(t)} e^{\frac{\langle \hat{\theta}_k(t) - \theta_k, Y_t \rangle + \gamma \|Y_t\|_{V_k^{-1}(t)} }{n_k(t)}} \sum_{s=1}^{t-1}  e^{\frac{\langle \theta_k - \hat{\theta}_k(s) , Y_s \rangle - \gamma \|Y_s\|_{V_k^{-1}(s)}}{e^{n_k(s)}}})
    \\
    = \mathbb{E}[R_k(t)] 
    ( 1 - \frac{1}{n_k(t)} e^{\frac{\langle \hat{\theta}_k(t) - \theta_k, Y_t \rangle + \gamma \|Y_t\|_{V_k^{-1}(t)}}{n_k(t)}} 
    \cdot \\
    \sum_{s=1}^{t-1}  e^{\frac{\langle \theta_k - \hat{\theta}_k(s), Y_s \rangle - \gamma \|Y_s\|_{V_k^{-1}(s)}}{e^{n_k(s)}}})
\end{multline}

\[
\|Y_t\| \leq 1
\]
\[
\|\theta_k\| \leq 1, \|\hat{\theta}_k(t)\| \leq 1
\]
Cauchy-Schwarz:
\[
|\langle Y_t, \theta_k \rangle| \leq \|Y_t\| \|\theta_k\| \leq 1
\]
\[
|\langle Y_t, \hat{\theta}_k(t) \rangle| \leq \|Y_t\| \|\hat{\theta}_k(t)\| \leq 1
\]

\[
e^{\frac{-2}{n_k(t)}} \leq e^{\frac{\langle Y_t, \theta_k - \hat{\theta}_k(t)\rangle}{n_k(t)}} \leq e^{\frac{2}{n_k(t)}}
\]

\begin{multline}
    1 - \frac{e^{\frac{2 + \gamma \|Y_t\|_{V_k^{-1}(t)}}{n_k(t)}}}{n_k(t)} \sum_{s=1}^{t-1} e^{\frac{2 - \gamma \|Y_s\|_{V_k^{-1}(s)}}{n_k(s)}} 
    \\
    \leq 1 - \frac{1}{n_k(t)}e^{\frac{\langle Y_t, \theta_k - \hat{\theta}_k(t)\rangle  + \gamma \|Y_t\|_{V_k^{-1}(t)}}{n_k(t)}}  
    \cdot \\
    \sum_{s=1}^{t-1} e^{\frac{\langle Y_s, \theta_k - \hat{\theta}_k(s)\rangle  - \gamma \|Y_s\|_{V_k^{-1}(s)}}{n_k(s)}} 
    \\
    \leq 1 - \frac{1}{n_k(t)}e^{\frac{-2 + \gamma \|Y_t\|_{V_k^{-1}(t)}}{n_k(t)}}\sum_{s=1}^{t-1} e^{\frac{-2 - \gamma \|Y_s\|_{V_k^{-1}(s)}}{n_k(s)}} 
\end{multline}

Denote: $\mathbb{E}[R_k(t)] = \lambda_k$.

\begin{multline} 
    \lambda_k \left( 1 - \frac{1}{n_k(t)}e^{\frac{2 + \gamma \|Y_t\|_{V_k^{-1}(t)}}{n_k(t)}} \sum_{s=1}^{t-1} e^{\frac{2  - \gamma \|Y_s\|_{V_k^{-1}(s)}}{n_k(s)}} \right) 
    \\
    \leq  \mathbb{E}[R_k(t)] - \mathbb{E}[G_k(t)] \leq
    \\
    \lambda_k \left( 1 - \frac{1}{n_k(t)}e^{\frac{-2  + \gamma \|Y_t\|_{V_k^{-1}(t)}}{n_k(t)}}\sum_{s=1}^{t-1} e^{\frac{-2 - \gamma \|Y_s\|_{V_k^{-1}(s)}}{n_k(s)}}  \right) 
\end{multline}


\textbf{The concentration of the remaining potential:}
\begin{theorem}\label{thm:concentration_rp}
    Concentration of the remaining potential with external factors:
    \begin{multline}
        \mathbb{E}\left[ e^{s(R_k(t)-\mathbb{E}[R_k(t)]} \right] 
        \\
        \leq
        \exp{\left( \frac{s^2 \lambda_k \alpha^2(\langle \theta_k, Y_t \rangle, n_k(t))}{4\sum_{l=1}^{t-1} \sum_{k^{'} \in I_l} \alpha(\langle\theta_{k^{'}},Y_l\rangle, n_{k^{'}}(l))} \right)}
    \end{multline}
\end{theorem}
\begin{proof}
    \begin{multline}
    \mathbb{E}\left[ e^{s(R_k(t)-\mathbb{E}[R_k(t)]} \right] =
    \\
    \Pi_{j=1}^{A_k} \left(1 - e^{-\lambda_j \sum_{l=1}^{t-1} \sum_{k^{'} \in I_l} \alpha(\langle \theta_{k^{'}}, Y_l\rangle, n_{k^{'}}(l))}\right)
    \cdot \\
    e^{-s\lambda_j\alpha(\langle \theta_k, Y_t\rangle, n_k(t))e^{-\lambda_j \sum_{l=1}^{t-1} \sum_{k^{'} \in I_l} \alpha(\langle \theta_{k^{'}}, Y_l\rangle, n_{k^{'}}(l))}} 
    + \\
    e^{-\lambda_j \sum_{l=1}^{t-1} \sum_{k^{'} \in I_l} \alpha(\langle \theta_{k^{'}}, Y_l\rangle, n_{k^{'}}(l))} 
    \cdot \\
    e^{s\lambda_j\alpha(\langle\theta_k,Y_t\rangle,n_k(t))\left(1-e^{-\lambda_j \sum_{l=1}^{t-1} \sum_{k^{'} \in I_l} \alpha(\langle \theta_{k^{'}}, Y_l\rangle, n_{k^{'}}(l))}\right)}
\end{multline}
    
By denoting, $p_j=e^{-\lambda_j \sum_{l=1}^{t-1} \sum_{k^{'} \in I_l} \alpha(\langle \theta_{k^{'}}, Y_l\rangle, n_{k^{'}}(l))}$ and \\ $t_j=s\lambda_j\alpha(\langle\theta_k,Y_t\rangle,n_k(t))$, we have, 

\begin{multline}
    \mathbb{E}\left[ e^{s(R_k(t)-\mathbb{E}[R_k(t)]} \right]
    = \\
    \Pi_{j=1}^{A_k} (1-p_j)e^{-p_j t_j} + p_j e^{t_j (1-p_j)}
    \\
    \leq \Pi_{j=1}^{A_k} exp{\left( \frac{1-2p_j}{4\ln((1-p_j)/p_j)} t_j^2 \right)} [\textbf{Theorem 3.2}\cite{berend2013concentration}]
    \\
    \leq \Pi_{j=1}^{A_k} exp{\left( \frac{1}{4\ln(1/p_j)} t_j^2 \right)} [\textbf{Lemma 3.5}\cite{berend2013concentration}]
    \\
    \leq \Pi_{j=1}^{A_k} exp{\left( \frac{s^2\lambda_j\alpha^2(\langle\theta_k,Y_t\rangle,n_k(t))}{4 \sum_{l=1}^{t-1} \sum_{k^{'} \in I_l} \alpha(\langle \theta_{k^{'}}, Y_l\rangle, n_{k^{'}}(l))}\right)}
    \\
    \leq \exp{\left( \frac{\alpha^2(\langle\theta_k,Y_t\rangle,n_k(t))}{4 \sum_{l=1}^{t-1} \sum_{k^{'} \in I_l} \alpha(\langle \theta_{k^{'}}, Y_l\rangle, n_{k^{'}}(l))} s^2\lambda_k \right)}
\end{multline}

Using Chernoff we finally have,
\begin{multline}
    P(R_k(t) > \mathbb{E}[R_k(t)]
    \\
    + \sqrt{\frac{ \alpha^2(\langle\theta_k,Y_t\rangle,n_k(t)) \lambda_k \ln(1/ \delta)}{\sum_{l=1}^{t-1} \sum_{k^{'} \in I_l} \alpha(\langle \theta_{k^{'}}, Y_l\rangle, n_{k^{'}}(l))}}) \leq \delta
\end{multline}
\end{proof}

\textbf{The concentration of the estimator of the remaining potential:}
\begin{equation}
    G_k(t) = \frac{1}{n_k(t)}\sum_{j=1}^{A_k} U_t^{\alpha} (j, k)
\end{equation}

For overlapping influencer supports, is the independence of the reward random variables preserved if L is constant per campaign and the reward is obtained by randomly dividing the feedback to each chosen influencer?

\begin{multline}
    U_t^{\alpha}(j) =
    \sum_{s=1}^{t-1} 
    \frac{\alpha(\langle\hat{\theta}_k(t), Y_t\rangle,n_k(t))}{\alpha(\langle\hat{\theta}_k(s), Y_s\rangle,n_k(s))}
    \cdot \\
    \mathbb{I}\{X_{s,j,k}=1,\{X_{l,j,k^{'}}=0\}_{k^{'}\in I_s\setminus \{k\}}, \{X_{l,j,k^{'}}=0\}_{l\neq s, k^{'}\in I_l}\}
    \\
    = \sum_{s=1}^{t-1} 
    e^{ \frac{\langle\hat{\theta}_k(t), Y_t\rangle + \gamma \|Y_t\|_{V_k^{-1}(t)}}{n_k(t)} - \frac{\langle\hat{\theta}_k(s), Y_s\rangle + \gamma \|Y_s\|_{V_k^{-1}(s)}}{n_k(s)}}
    \cdot \\
    \mathbb{I}\{X_{s,j,k}=1,\{X_{l,j,k^{'}}=0\}_{k^{'}\in I_s\setminus \{k\}}, \{X_{l,j,k^{'}}=0\}_{l\neq s, k^{'}\in I_l}\}
\end{multline}

\begin{multline}
    G_k(t) = 
    \frac{1}{n_k(t)} \sum_{s=1}^{t-1} 
    \cdot \\
    \mathbb{I} \left\{X_{s,j,k}=1, \{X_{l,j, k^{'}}=0\}_{l \neq s, k^{'} \in I_l}, \{X_{s,j, k^{'}}=0\}_{k^{'} \in I_s \setminus \{k\}}\right\}
    \cdot \\
    e^{ \frac{\langle\hat{\theta}_k(t), Y_t\rangle  + \gamma \|Y_t\|_{V_k^{-1}(t)}}{n_k(t)} - \frac{\langle\hat{\theta}_k(s), Y_s\rangle  + \gamma \|Y_s\|_{V_k^{-1}(s)}}{n_k(s)}}
\end{multline}

For Bennett's inequality:

Denote $X_t(j) = \frac{U_t^{\alpha}(j)}{n_k(t)}$.

Denote $X_t(j,k) = \frac{U_t^{\alpha}(j,k)}{n_k(t)}$.

\begin{multline}
    X_t(j) =  \sum_{s=1}^{t-1} e^{ \frac{\langle\hat{\theta}_k(t), Y_t\rangle  + \gamma \|Y_t\|_{V_k^{-1}(t)}}{n_k(t)}  - \frac{\langle\hat{\theta}_k(s), Y_s\rangle  + \gamma \|Y_s\|_{V_k^{-1}(s)}}{n_k(s)}} 
    \cdot \\
    \frac{\mathbb{I} \left\{X_{s,j,k}=1, \{X_{l,j, k^{'}}=0\}_{l \neq s, k^{'} \in I_l}, \{X_{s,j, k^{'}}=0\}_{k^{'} \in I_s \setminus \{k\}}\right\} }{n_k(t)}
    \\
    \leq \frac{1}{n_k(t)} e^{ \frac{1 + \gamma \|Y_t\|_{V_k^{-1}(t)}}{n_k(t)}}\sum_{s=1}^{t-1}e^{ \frac{1 + \gamma \|Y_s\|_{V_k^{-1}(s)}}{n_k(s)}}
\end{multline}

\begin{equation}
    X_t(j,k) \leq 
    \frac{1}{n_k(t)} e^{ \frac{1 + \gamma \|Y_t\|_{V_k^{-1}(t)}}{n_k(t)}}\sum_{s=1}^{t-1}e^{ \frac{1  + \gamma \|Y_s\|_{V_k^{-1}(s)}}{n_k(s)}}
\end{equation}

\begin{equation}
    S = \sum_{j=1}^{A_k} (X_t(j) - \mathbb{E}[X_t(j)])
\end{equation}

\begin{multline}
    v = \sum_{j=1}^{A_k} \mathbb{E}[X_t^2(j)]
    = \sum_{j=1}^{A_k} \mathbb{E}\left[\frac{U_t^{\alpha}(j)^2}{n_k(t)^2}\right]
    \\
    = \frac{\sum_{j=1}^{A_k} \mathbb{E}[U_t^{\alpha}(j)^2]}{n^2_k(t)}
    = \sum_{j=1}^{A_k} \sum_{s=1}^{t-1} \frac{1^2 P(C_j(t-1) = 1)}{n^2_k(t)}
    \cdot \\
    e^{ \frac{2\langle\hat{\theta}_k(t), Y_t\rangle + 2\langle \gamma \|Y_t\|_{V_k^{-1}(t)}, Y_t \rangle}{n_k(t)} - \frac{2\langle\hat{\theta}_k(s), Y_s\rangle  + 2\langle \gamma \|Y_s\|_{V_k^{-1}(s)}, Y_s \rangle}{n_k(s)}}
    \\
    = \sum_{j=1}^{A_k} \sum_{s=1}^{t-1}  \frac{\lambda_j }{n^2_k(t)} 
    e^{-\lambda_j\sum_{l=1}^{t-1} \sum_{k^{'} \in I_l} e^{\frac{\langle\theta_{k^{'}}, Y_l\rangle}{n_{k^{'}}(l)}}}
    \cdot \\
    e^{\frac{\langle\theta_k, Y_t\rangle}{n_k(t)} +
    \frac{2\langle\hat{\theta}_k(t),Y_t\rangle  + 2\gamma \|Y_t\|_{V_k^{-1}(t)}}{n_k(t)}
    +
    \frac{-2\langle\hat{\theta}_k(s),Y_s\rangle - 2\gamma \|Y_s\|_{V_k^{-1}(s)}}{n_k(s)}}
\end{multline}

\begin{multline} 
    S = \frac{1}{n^2_k(t)}
    \cdot \\
    \sum_{j=1}^{A_k} \sum_{s=1}^{t-1}  
    e^{\frac{\langle\theta_k, Y_t\rangle}{n_k(t)}}
    \lambda_j
    e^{-\lambda_j\sum_{l=1}^{t-1} \sum_{k^{'} \in I_l}
    e^{\frac{\langle\theta_{k^{'}}, Y_l\rangle}{n_{k^{'}}(l)}}}
    \cdot \\
    e^{\frac{2\langle\hat{\theta}_k(t),Y_t\rangle + 2\gamma \|Y_t\|_{V_k^{-1}(t)}}{n_k(t)}} e^{\frac{-2\langle\hat{\theta}_k(s),Y_s\rangle - 2\gamma \|Y_s\|_{V_k^{-1}(s)}}{n_k(s)}}
\end{multline}

\begin{equation}
v \leq  \frac{e^{\frac{3 + 2\gamma \|Y_t\|_{V_k^{-1}(t)}}{n_k(t)}} \sum_{s=1}^{t-1} e^{ \frac{2 - 2\gamma \|Y_s\|_{V_k^{-1}(s)}}{n_k(s)}}}{n^2_k(t)} \sum_{j=1}^{A_k} \lambda_j
\end{equation}

\begin{equation}
\sum_{j=1}^{A_k} \lambda_j \leq \lambda_k << A_k
\end{equation}

\begin{equation}
v \leq  \frac{e^{\frac{3 + 2\gamma \|Y_t\|_{V_k^{-1}(t)}}{n_k(t)}} \sum_{s=1}^{t-1} e^{ \frac{2 - 2\gamma \|Y_s\|_{V_k^{-1}(s)}}{n_k(s)}}}{n^2_k(t)}  \lambda_k 
\end{equation}

\begin{multline}
P(G_k(t) - \mathbb{E}[G_k(t)] > 
\\
\sqrt{\frac{2\lambda_k e^{\frac{3 + 2\gamma \|Y_t\|_{V_k^{-1}(t)}}{n_k(t)}} \sum_{s=1}^{t-1} e^{ \frac{2 - 2\gamma \|Y_s\|_{V_k^{-1}(s)}}{n_k(s)}}}{n^2_k(t)}\ln \frac{1}{\delta}} +
\\
\frac{e^{ \frac{1 + \gamma \|Y_t\|_{V_k^{-1}(t)}}{n_k(t)}}\sum_{s=1}^{t-1}e^{ \frac{1 + \gamma \|Y_s\|_{V_k^{-1}(s)}}{n_k(s)}}}{3n_k(t)} \ln\frac{1}{\delta}) \leq \delta
\end{multline}

\textbf{The confidence interval:}
\begin{multline} R_k(t) - G_k(t) \leq \sqrt{\frac{ \alpha^2(\langle\theta_k,Y_t\rangle,n_k(t)) \lambda_k \ln(1/ \delta)}{\sum_{l=1}^{t-1} \sum_{k^{'} \in I_l} \alpha(\langle \theta_{k^{'}}, Y_l\rangle, n_{k^{'}}(l))}}
\\
+ \sqrt{\frac{2\lambda_k e^{\frac{3+ 2\gamma \|Y_t\|_{V_k^{-1}(t)}}{n_k(t)}} \sum_{s=1}^{t-1} e^{ \frac{2- 2\gamma \|Y_s\|_{V_k^{-1}(s)}}{n_k(s)}}}{n^2_k(t)}\ln \frac{1}{\delta}}
\\
+ \frac{e^{ \frac{1+ \gamma \|Y_t\|_{V_k^{-1}(t)}}{n_k(t)}}\sum_{s=1}^{t-1}e^{ \frac{1 + \gamma \|Y_s\|_{V_k^{-1}(s)}}{n_k(s)}}}{3n_k(t)} \ln\frac{1}{\delta} 
\\
+ \lambda_k \left( 1 - e^{\frac{-2 + \gamma \|Y_t\|_{V_k^{-1}(t)}}{n_k(t)}} \frac{1}{n_k(t)} \sum_{s=1}^{t-1} e^{\frac{-2 - \gamma \|Y_s\|_{V_k^{-1}(s)}}{n_k(s)}} \right).
\end{multline}

\begin{multline}
-\beta_k(t) + 
\Omega\left(\frac{T\lambda_k(T)}{n_k(T)}e^{\frac{\gamma \|Y_T\|_{V_k^{-1}(T)}}{n_k(T)}}\right)
\leq \\
R_k(t) - G_k(t) \leq \beta_k(t) +
\mathcal{O}\left(T\frac{\lambda_k(T)}{n_k(T)}e^{\frac{\gamma \|Y_T\|_{V_k^{-1}(T)}}{n_k(T)}}\right),
\end{multline}
where:
\[
\beta_k(t) = 
\sqrt{\frac{2\lambda_k(t) e^{\frac{3+ 2\gamma \|Y_t\|_{V_k^{-1}(t)}}{n_k(t)}} \sum_{s=1}^{t-1} e^{ \frac{2 - 2\gamma \|Y_s\|_{V_k^{-1}(s)}}{n_k(s)}}}{n^2_k(t)}\ln \frac{1}{\delta}}
 + \]
 \[
\sqrt{\frac{ e^{2/n_k(t)} \lambda_k(t) \ln(1/ \delta)}{\sum_{s=1}^{t-1} \sum_{k^{'} \in I_s} e^{-1/n_{k^{'}}(s)}}} +
\]\[
 \frac{e^{ \frac{1+ \gamma \|Y_t\|_{V_k^{-1}(t)}}{n_k(t)}}\sum_{s=1}^{t-1}e^{ \frac{1+ \gamma \|Y_s\|_{V_k^{-1}(s)}}{n_k(s)}}}{3n_k(t)} \ln\frac{1}{\delta}
\]
\end{proof}

\begin{table}[h]
    \centering
    \caption{Data statistics.}
    \begin{tabular}{ccc}
        \toprule
       Net. & \# Users & \#Edges \\
        \midrule
       Weibo & 1,776,950 & 308M \\
       Twitter & 11.6M & 309M \\
       \bottomrule
    \end{tabular}
    \begin{tabular}{ccc}
        \toprule
       Net. & \#Orig.-microblogs & \#Retweets \\
        \midrule
       Weibo & 300,000 & 23,755,810 \\
       Twitter & 242M & 341,811,085 \\
       \bottomrule
    \end{tabular}
    \label{tab:ds}
\end{table}

\section{Twitter dataset statistics}

\begin{table}[h]
    \centering
    \begin{tabular}{c|c|c}
      & Random tweets   & Random campaign \\
      count & 705586 & 705586 \\ 
      mean & 0.6 & 3.5 \\
      std & 6.68 & 16.37 \\
      min & 0 & 0 \\
      25\% & 0 & 1 \\
      50\% & 0 & 2 \\
      75\% & 1 & 4 \\
      max & 4581 & 10676 \\
    \end{tabular}
    \caption{The rewards distribution in the datasets.}
    \label{tab:rewards}
\end{table}

The statistics for both types of simulation are presented in Table~\ref{tab:rewards}.


\end{document}